\documentclass{article}

\usepackage{arxiv}

\usepackage{times}  
\usepackage{helvet}  
\usepackage{courier}  
\usepackage{url}  
\usepackage{graphicx}  
\usepackage{amsmath}
\usepackage{multirow}
\usepackage{multicol}
\usepackage{makecell}
\usepackage{amsfonts}
\usepackage{mathtools}
\usepackage{color}
\usepackage{nameref}
\usepackage{tabularx}

\newcommand\blfootnote[1]{%
  \begingroup
  \renewcommand\thefootnote{}\footnote{#1}%
  \addtocounter{footnote}{-1}%
  \endgroup
}

\newcommand{\vectorproj}[2][]{\textit{proj}_{{#1}}{#2}}
\newcolumntype{Y}{>{\centering\arraybackslash}X}

\DeclarePairedDelimiter\floor{\lfloor}{\rfloor}

\begin{document}

\title{CircConv: A Structured Convolution with Low Complexity}

\author{Siyu Liao\textsuperscript{1*},
Zhe Li\textsuperscript{2*},
Liang Zhao\textsuperscript{4*}, 
Qinru Qiu\textsuperscript{3},
Yanzhi Wang\textsuperscript{5},
Bo Yuan\textsuperscript{1}
\\
\textsuperscript{1}{Department of Electrical and Computer Engineering, Rutgers University}\\
\textsuperscript{2}{Google AI}\\
\textsuperscript{3}{Electrical Engineering and Computer Science Department, Syracuse University}\\
\textsuperscript{4}{Department of Computer Science, City University of New York}\\
\textsuperscript{5}{Department of Electrical and Computer Engineering, Northeastern University}\\
siyu.liao@rutgers.edu,
lzhe@google.com,
liang.zhao1@lehman.cuny.edu,
qiqiu@syr.edu,\\
yanz.wang@northeastern.edu,
bo.yuan@soe.rutgers.edu
}

\maketitle

\blfootnote{*Siyu Liao, Zhe Li and Liang Zhao contributes equally to this work. Work was done when Zhe Li was with Syracuse University.}

\begin{abstract}
Deep neural networks (DNNs), especially deep convolutional neural networks (CNNs), have emerged as the powerful technique in various machine learning applications. However, the large model sizes of DNNs yield high demands on computation resource and weight storage, thereby limiting the practical deployment of DNNs. To overcome these limitations, this paper proposes to impose the circulant structure to the construction of convolutional layers, and hence leads to circulant convolutional layers (CircConvs) and circulant CNNs. The circulant structure and models can be either trained from scratch or re-trained from a pre-trained non-circulant model, thereby making it very flexible for different training environments.  Through extensive experiments, such strong structure-imposing approach is proved to be able to substantially reduce the number of parameters of convolutional layers and enable significant saving of computational cost by using fast multiplication of the circulant tensor.
\end{abstract}

\section{Introduction} \label{sec:intro}
Large-scale deep neural networks (DNNs), especially deep convolutional neural networks (CNNs), have achieved extraordinary success in various artificial intelligence applications such as image recognition, video analysis, etc. \cite{krizhevsky2012imagenet,he2016deep,karpathy2014large}. However, the large model sizes of DNNs make themselves both computation-intensive and memory-intensive, thereby potentially hindering the expected widespread deployment of DNNs in many latency-sensitive resource-constrained applications.

To address these limitations, many approaches \cite{han2015deep,gong2014compressing,wen2016learning,feng2015learning} have been proposed to reduce the computational cost and/or memory footprint of DNNs. In general, those existing efforts can be roughly categorized as two types: fully-connected layer-oriented reduction, such as connection pruning \cite{han2015deep}\footnote{It can also bring reduction for convolutional layers to some degree. But the most reduction in the number of parameters is achieved on fully-connected layers.}, weight clustering \cite{gong2014compressing}, and convolutional layer-oriented reduction, such as low rank approximation \cite{jaderberg2014speeding}, sparsity regularization \cite{wen2016learning,feng2015learning}. Nowadays, consider 1) convolutional layers consume most of the computational processing in DNNs and 2) many state-of-the-art DNNs, such as ResNet \cite{he2016deep} and Inception \cite{szegedy2015going}, use very few fully-connected layers that only contain a small portion of parameters of the entire models (e.g. less than 5$\%$ parameters for fully-connected layers in ResNet-152), the reduction on computational cost and numbers of parameters of convolutional layers become very essential.

\begin{figure}[!tb]
\centering
\includegraphics[width = 0.6\columnwidth]{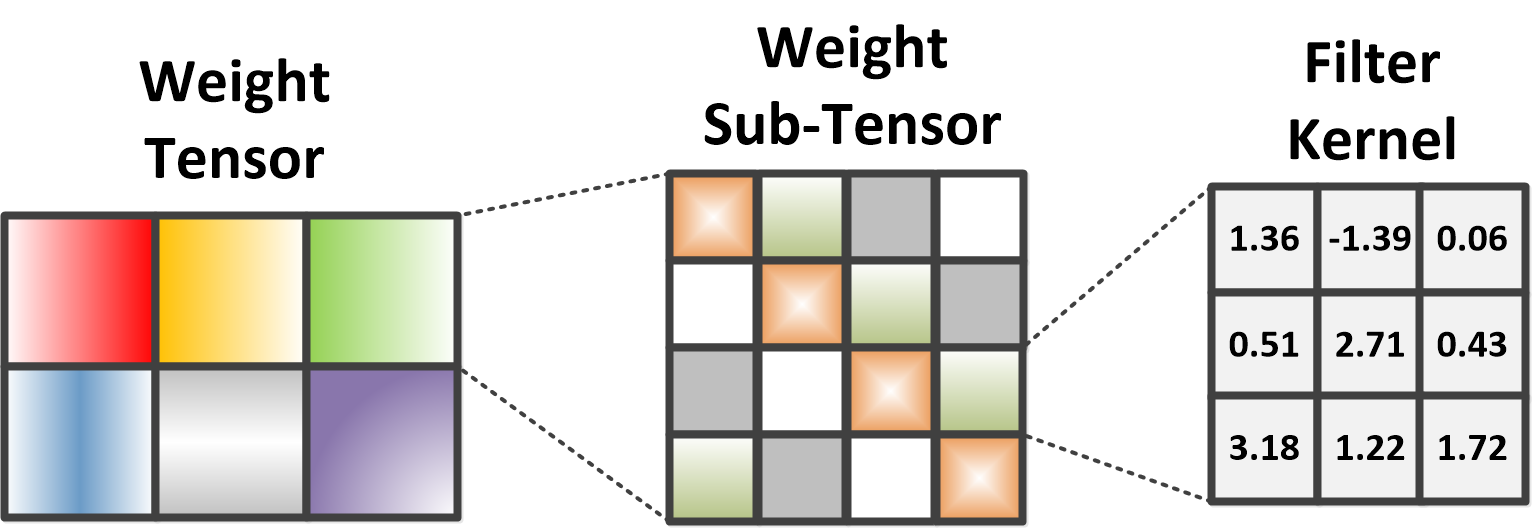}
\caption{Illustration of a circulant weight tensor. Blocks of the same color in the middle share the same set of kernel weights (on the right). This significantly reduces the total amount of parameters needed to represent this tensor. In addition, the placement of blocks displays a circulant structure, facilitating FFT-based fast algorithms.}
\label{fig:tensor}
\end{figure}

\textbf{Technical preview and advantages.}
In this paper we propose to impose the circulant structure to the construction of convolutional layers shown in Figure \ref{fig:tensor}, yielding low-computation-complexity, low-space-cost \emph{circulant convolutional layers (CircConvs)} and the corresponding \emph{circulant CNNs}. Different from prior convolutional layer-oriented compression approaches that are based on the unstructured tensors, the model-size reduction in this paper results from the use of \emph{circulant tensors} \cite{rezghi2011diagonalization}: The weight tensors for convolutional layers, which were original unstructured, are now constructed in the circulant format, thereby leading to substantial reduction in computational cost and numbers of parameters. In short, the proposed approach brings the following advantages:

\textit{1)} It reduces the space cost of the convolutional layers because of the inherent spatial regularity of circulant tensors, thereby resulting in high compression ratios for the overall network model sizes. 

\textit{2)} It saves the computation of the convolutional layers by leveraging the fast circulant tensor-specific multiplication algorithm, and hence greatly reduces the computational cost of the entire networks. 

\textit{3)} It enables the improved accuracy for the corresponding circulant CNNs as compared with the similar-size non-circulant CNNs. In other words, the benefits of model-size reduction resulting from using circulant convolutional layers can translate to the increase of accuracy. 

\textit{4)} The circulant structure can be imposed by either training from scratch or re-training from a pre-trained non-circulant model, thereby making circulant CNNs very flexible for different training environments.

We conduct extensive experiments on circulant CNNs and the results show that the proposed circulant structure-imposing approach can effectively reduce the model sizes and floating point operations (FLOPs) with negligible accuracy drop. In addition, the experiments on wide ResNets show that the proposed approach leads to better accuracy than the non-circulant ResNet models with similar numbers of parameters. Furthermore, we also compare the 
FLOPs of the proposed circulant CNN models with the non-circulant CNN models, and experimental results show that our proposed method can reduce FLOPs for the inference. 

\section{Related Work} \label{sec:related}
\textbf{Weight pruning/clustering.} \cite{gong2014compressing} proposes to cluster the weights to reduce the model sizes of DNNs. In that work, various types of weight clustering, including scalar quantization, product quantization, residual quantization, are investigated. 
\cite{han2015deep} proposes a multi-step compression pipeline comprising of weight pruning, clustering, and quantization to achieve high compression ratios for the entire networks. Notice that as indicated in \cite{wen2016learning}, because most parameter reduction in \cite{gong2014compressing} and \cite{han2015deep} are achieved on fully-connected layers, the reduction in the computational cost of convolutional layers is not significant. 
In addition, it is also found that weight parameters in frequency domain can be pruned \cite{wang2016cnnpack} or clustered via a hashing function \cite{chen2016compressing}.

\textbf{Low rank approximation (LRA).} LRA is an efficient approach to compress DNNs \cite{jaderberg2014speeding,sainath2013low,zhao2016low}. In \cite{jaderberg2014speeding}, various types of LRA-based solutions are proposed to reduce the numbers of parameters and computational cost of convolutional layers. However, the model-size reduction using LRA usually requires costly reiterations of decomposition and fine-tuning to minimize the approximation error and retain the accuracy.

\textbf{Sparsity regularization.} Increasing the sparsity of network by performing regularization is another popular technique to reduce model sizes of DNNs. \cite{feng2015learning}, \cite{girosi1995regularization} and \cite{wen2016learning} propose several sparsity-introducing techniques by leveraging different types of regularization, such as L1-norm, group-lasso etc. Though sparsity regularization essentially provides a stable reduction in computational cost, the resulting reduction in model size is not significant.

\textbf{Structured transform.} By using structured matrices, the structured transform can enable very high compression ratios for fully-connected (FC) layers. In \cite{sindhwani2015structured,cheng2015exploration,moczulski2015acdc}, weight matrices are constructed in the format of structured matrices to achieve significant reduction in model sizes.
\cite{zhao2017theoretical} further proves that the low displacement rank-based neural networks, which are the generalization of the structured networks can still exhibit universal approximation property. However, the FC layer-specific approaches in  \cite{sindhwani2015structured,cheng2015exploration,moczulski2015acdc} cannot be directly applied to the popular and important convolutional layers. Instead, consider FC layer can be viewed as a type of special convolutional layer, our proposed circulant convolution-imposing approach has more generality and is very useful for practical applications. 
Particularly, compared with \cite{cheng2015exploration} and \cite{sindhwani2015structured}, we generalize the structure from regular 2D weight matrix in fully connected layer to the 4D weight tensor in convolutional layer, where the underlying computation is totally different. 

Moreover, this paper is significantly different from two related works \cite{ding2017c,yanzhi2018} in following aspects:
1) we propose an approach on direct operation on 4D circulant tensor with all the algorithm-level details; while prior works require extra processing step to convert tensor to matrix; 
2) we perform comprehensive experiments on circulant convolution and present accuracy results and analysis on different datasets;
3) we propose a novel algorithm to convert non-circulant tensor into circulant tensor, thereby making obtaining circulant convolution on pre-trained non-circulant models become possible; while prior works can only train the structure from scratch.

\section{Imposing Circulant Structure to Convolutional Layers} \label{sec:circconv}
\subsection{Circulant Convolutional Layer}\label{sec:circlayer}

\textbf{Conventional convolutional layer.} 
In general, a convolutional layer maps a 3-dimensional input tensor $\mathcal{X} \in \mathbb{R}^{W_0 \times H_0 \times C_0}$ into a 3-dimensional output tensor $\mathcal{Y} \in \mathbb{R}^{W_2 \times H_2 \times C_2}$ through convolution with a 4-dimensional kernel tensor
$\mathcal{W}\in\mathbb{R}^{W_1\times H_1\times C_0\times C_2}$. Here $W_i$ and $H_i$ for $i=0,1,2$, are the spatial width and height of the input, kernel, and output tensor, respectively; $C_0$ and $C_2$ are the number of input channels and output channels. The convolution operation is expressed as follows:

\begin{equation}\label{eqn:conv}
\mathcal{Y}(w_2, h_2, c_2) = \sum_{w_1=1}^{W_1} \sum_{h_1=1}^{H_1} \sum_{c_0=1}^{C_0}  \Big(\mathcal{X} (w_2-w_1, h_2-h_1, \\c_0) \cdot \mathcal{W} (w_1, h_1, c_0, c_2)\Big).
\end{equation}

Although stride can be set for convolution, we consider the case of stride that equals to 1 to make a better understanding of circulant convolution. It should be noted that stride wouldn't affect our convolution algorithm and design. Moreover, we can express Eq. \ref{eqn:conv} in the form of a fiber multiplied by a slice as below:

\begin{equation}\label{eqn:conv2}
\mathcal{Y}(w_2, h_2, :) = \sum_{w_1=1}^{W_1} \sum_{h_1=1}^{H_1}  \Big(\mathcal{X} (w_2-w_1, h_2-h_1, :) \\
\ast \mathcal{W} (w_1, h_1, :, :)\Big),
\end{equation}
where $*$ and $:$ denote the matrix-vector multiplication and the range of indices, respectively.

\textbf{Circulant convolutional layer.} Different from a conventional convolutional layer, the circulant convolutional layer has a weight tensor $\mathcal{W}$ that exhibits circulant structure. In other words, the $\mathcal{W}$ of a circulant convolution layer is a 4D \emph{circulant tensors} \cite{rezghi2011diagonalization}. In general, a circulant tensor can exhibit circulant structure along any pair of its dimensions. However, as 
$W_1$ and $H_1$ are usually much smaller than $C_0$ and $C_2$ for tensor $\mathcal{W}$, we impose the circulant structure along the input channel and output channel dimensions to achieve high model-size compression ratio. Note that in practice we need to partition the tensor $\mathcal{W}$ into circulant sub-tensors of size $W_1\times H_1\times N\times N$. This is necessary because the circulant structure requires that the two corresponding dimension must be equal, while $C_0$ and $C_2$ are usually not the same. Larger $N$ means larger compression ratio but it could hurt the model performance to some degree. 
By adjusting the partition size $N$ we can balance the trade-off between compression ratio and model accuracy. 

More specifically, let $N$ be the partition size with $C_0 = R\times N$ and $C_2 = S\times N$ \footnote{Zero-padding is needed when $N$ does not divide $C_0$ or $C_2$.}, then $\mathcal{W}$ can be defined by a 4-dimensional base tensor $\mathcal{W'} \in \mathbb{R}^{W_1 \times H_1 \times RN \times S}$:
\begin{equation}\label{eqn:tensor}
\textstyle \mathcal{W} (w_1, h_1, c_0, c_2) = \mathcal{W'} (w_1, h_1, p, q),
\end{equation}
where $p,q$ are indices satisfying $\floor{c_0/N}=\floor{p/N}$, $\floor{c_2/N}=q$, and $c_0 - c_2\equiv p\pmod{N}$.
Fig. \ref{fig:tensor} illustrates the circulant structure of weight tensor $\mathcal{W}$. From this figure, it can be seen that the circulant structure is imposed to $\mathcal{W}$ along the input/output channel dimensions. The block-circulant weight tensor consists of six circulant weight sub-tensors, where different colors represent different circulant weight sub-tensors. Each circulant weight sub-tensor consists of sixteen kernel filters that are represented in  different colors such as green and yellow.

\subsection{Fast Forward and Backward Propagation Schemes on Circulant Convolutional Layer} \label{subsec:fpbp}

Eq. \ref{eqn:tensor} shows that the weight tensor $\mathcal{W}$ of a circulant convolutional layer exhibits the circulant structure and has the reduced number of independent parameters. Besides, according to the tensor theory \cite{rezghi2011diagonalization}, circulant tensor also has the advantage of fast multiplication. Since multiplication is the kernel computation in neural network training and inference, the existence of fast multiplication of circulant tensor enables the immediate reduction in computational cost. Next, we describe the fast forward and backward propagation schemes by leveraging the fast multiplication of circulant weight tensor.

\textbf{Fast forward propagation.} We first present the fast forward propagation scheme. Recall that Eq. \ref{eqn:conv2} is the forward propagation scheme for a general convolutional layer. To ease the notation, define $N_k = ((k-1)N+1,...,kN)$ for $k=1,...,\max(R,S)$, and rewrite Eq. \ref{eqn:conv2} as below:

\begin{equation}\label{eqn:conv3}
\mathcal{Y}(w_2, h_2, N_i) = \sum_{w_1=1}^{W_1} \sum_{h_1=1}^{H_1} \sum_{j=1}^{R} \Big(\mathcal{X} (w_2-w_1, \\ h_2-h_1, N_j) 
\ast \mathcal{W} (w_1, h_1, N_j,N_i)\Big),
\end{equation}
where $i \in \{1, \dots, S\}$. According to \cite{rezghi2011diagonalization,pan2012structured}, Fast Fourier Transform (FFT)  can be used to accelerate the multiplication of a fiber and a slice of circulant tensor with time complexity reduced from $O(N^2)$ to $O(N\log N)$. Therefore, when $\mathcal{W}$ is a circulant tensor, Eq. \ref{eqn:conv3} can be reformulated using FFT as below:

\begin{equation}\label{eqn:conv4}
\mathcal{Y}(w_2, h_2, N_i)  
 = \mathrm{ifft}\Big( \sum_{w_1=1}^{W_1} \sum_{h_1=1}^{H_1}\sum_{j=1}^{R} \mathrm{fft}\big(\mathcal{X} (w_2-w_1,
 \\ h_2-h_1, N_j)\big) 
    \circ \mathrm{fft}\big(\mathcal{W'} (w_1, h_1, N_j, N_i)\big)\Big).
\end{equation}
Here $\circ$ is the element-wise multiplication.

\textbf{Fast backward propagation.} 
Now consider backward propagation. Given loss function $L$, it is well known that the goal of backpropgation algorithm \cite{lecun1998gradient} is to compute gradients of loss function $L$ with respect to each weight and input. 
Hence according to the chain rule, the gradient computation for circulant convolutional layer can be derived from Eq. \ref{eqn:tensor} and Eq. \ref{eqn:conv3} as below:

\begin{equation}\label{eqn:gradient}
    \frac{\partial L}{\partial \mathcal{W'} (w_1, h_1, p, q)} 
= \sum_{w_2=1}^{W_2} \sum_{h_2=1}^{H_2} \sum_{c_2=(q-1)N+1}^{qN}
\\
\frac{\partial L}{\partial \mathcal{Y}(w_2, h_2, c_2)} \frac{\partial \mathcal{Y}(w_2, h_2, c_2)}{\partial \mathcal{W'} (w_1, h_1, p, q)},
\end{equation}
\begin{equation}\label{eqn:gradient2}
\frac{\partial L}{\partial \mathcal{X} (x, y, c_0)} = \sum_{w_1=1}^{W_1} \sum_{h_1=1}^{H_1} \sum_{c_2\equiv c_0(\textit{ mod }N)} \\ 
\frac{\partial L}{\partial \mathcal{Y}(w_1+x, h_1+y, c_2)} \frac{\partial \mathcal{Y}(w_1+x, h_1+y, c_2)}{\partial \mathcal{X} (x, y, c_0)}.
\end{equation}

Again, according to \cite{rezghi2011diagonalization}, when $\mathcal{W}$ is a circulant tensor, Eq. \ref{eqn:gradient} and Eq. \ref{eqn:gradient2} can also be accelerated by using FFT as below:

\begin{equation}\label{eqn:gradient3}
\frac{\partial L}{\partial \mathcal{W'}(w_1, h_1, N_j, i)}
 = \text{ifft}( \sum_{w_2=1}^{W_2} \sum_{h_2=1}^{H_2} \\ \text{fft}(\frac{\partial L}{\partial \mathcal{Y}(w_2, h_2, N_i)}) \circ \text{fft}(\mathbf{x'_j}))
\end{equation}
\begin{equation}\label{eqn:gradient4}
\frac{\partial L}{\partial \mathcal{X} (x, y, N_j)} 
= \text{ifft}( \sum_{w_1=1}^{W_1} \sum_{h_1=1}^{H_1} 
\sum_{i=1}^{S} 
\text{fft}( 
\\
\frac{\partial L}{\partial \mathcal{Y}(w_1+x, h_1+y, N_i)})
\circ
\text{fft}(\mathbf{w'_{j,i}})
),
\end{equation}
where $\mathbf{x'_j}$ and $\mathbf{w'_{j,i}}$ are fibers $\mathcal{X}(w_1, h_1,T)$ and $\mathcal{W}'(x,y, (j-1)N+T,i)$ with $T=(1, N, ..., 2)$.

\textbf{Capability of training Circulant CNN from scratch.} It should be noted that the gradient computations described in Eq. \ref{eqn:gradient3} and Eq. \ref{eqn:gradient4} are actually based on $\mathcal{W'}$. Since we can always construct the circulant tensor $\mathcal{W}$ from base tensor $\mathcal{W'}$ using Eq. \ref{eqn:tensor}, Eq. \ref{eqn:gradient3} and \ref{eqn:gradient4} imply that the circulant structure of weight tensor $\mathcal{W}$ is always kept during the training phase. In other words, if we initialize $\mathcal{W}$ as the circulant tensor at the initialization stage of training, then during the training procedure Eq. \ref{eqn:gradient3} and \ref{eqn:gradient4} can guarantee $\mathcal{W}$ always exhibit circulant structure. Therefore, a circulant CNN can be completely trained from the scratch. 

\subsection{Conversion from Non-circulant Tensor to Circulant Tensor}

Forward and backward propagation section
indicates that a circulant convolutional layer can be trained from scratch. In this subsection, we also present a conversion technique that can directly convert a non-circulant weight tensor to a circulant one. Such conversion is very useful when a pre-trained model is already available and needs to be imposed with circulant structure. 

Specifically, the proposed conversion technique is based on the circulant approximation approach \cite{chu2003real} used for circulant matrix. In matrix theory, let $\mathbf{Z_1}  \in \mathbb{R}^{N\times N} $ denote a permutation matrix as following:
\begin{equation}
    \mathbf{Z_1} = 
    \begin{bmatrix}
    0 & 1 & 0 & \dots  & 0 \\
    0 & 0 & 1 & \dots  & 0 \\
    \vdots &  & \ddots & \ddots & \vdots \\
    0 & & & & 1 \\
    1 & 0 & 0 & \dots  & 0
\end{bmatrix}.
\end{equation}
Then a circulant matrix $\mathbf{W}_{circ} \in \mathbb{R}^{N\times N} $ with its first row  $\mathbf{w}=(w_0, w_1, \dots, w_{N-1})$ can be represented in the polynomial form of $\mathbf{Z_1}$ as follows:
\begin{equation}
    \textstyle\mathbf{W}_{circ} = \sum_{i=0}^{N-1} w_i \mathbf{Z_1}^i.
\end{equation}

According to \cite{chu2003real}, for a non-circulant matrix $\mathbf{W}_{non-circ} \in \mathbb{R}^{N\times N}$, its nearest circulant matrix $\mathbf{W}_{circ}$ (measured in the Frobenius norm) is given by projection:
\begin{equation}\label{eqn:cirappro}
\begin{split}
 \mathbf{w} &= \vectorproj[N]{\mathbf{W}_{non-circ}},
\quad \\
\forall w_i \in \mathbf{w},  
    w_i &= \frac{1}{N} \langle \mathbf{W}_{non-circ}, \mathbf{Z_1}^i \rangle_\mathbf{F},
\end{split}
\end{equation}
where $\langle\cdot,\cdot\rangle_\mathbf{F}$ is the Frobenius inner product. 

Note that the 4-D weight tensor of a convolutional layer can be viewed as a matrix of size $W_1\times W_2$ where each entry is a matrix of size $C_0\times C_2$. Therefore, by using Eq. \ref{eqn:cirappro}, the conversion from a non-circulant tensor $\mathcal{W}_{non-circ}$
to a circulant tensor $\mathcal{W}_{circ}$ can be achieved by performing the projection as follows:
\begin{equation}
    \textstyle \mathcal{W'}(w_1, h_1, N_j, i) = \vectorproj[N]{\mathcal{W}_{non-circ}(w_1, h_1, N_j, N_i)},
    \label{eqn:circ_approx}
\end{equation}
where $\mathcal{W'}$ is the base tensor that defines circulant tensor $\mathcal{W}_{circ}$, and the mapping from $\mathcal{W'}$ to $\mathcal{W}_{circ}$ is given in Eq. \ref{eqn:tensor}.

\textbf{Capability of training Circulant CNN from a pre-trained model.} Based on the conversion scheme shown in Eq. \ref{eqn:circ_approx}, any non-circulant convolutional layer of a pre-trained model can be directly converted to a circulant convolutional layer. Typically such direct conversion brings non-negligible accuracy drop incurred by the approximation error. In order to recover the accuracy, further re-training on the converted model is needed by following the backward propagation scheme in Eq. \ref{eqn:gradient3} and \ref{eqn:gradient4}. Consequently, a non-circulant pre-trained model can be imposed with circulant structure by using the proposed circulant conversion and re-training schemes with preserving high accuracy.

\subsection{Efficiency on Space and Computation}
Table \ref{tbl:fftcomp} summarizes the space and time complexity of the circulant convolutional layers. 
It can be seen that the proposed circulant structure-imposing approach enables simultaneous improvement on both space efficiency and computation efficiency. 
Larger $N$ can result in larger FFT size and lower space and time complexity.
Also, compared with the 2-D FFT-based fast convolution in \cite{mathieu2013fast}, our 1-D FFT-based approach has much lower space and time complexity since $N$ is typically much larger than $R$ and $S$. 
\begin{table}[!tb]
\centering
\caption{Comparison with \cite{mathieu2013fast} in terms of FFT, time and space complexity, where $N=C_0=C_2$.
}
\label{tbl:fftcomp}
\begin{tabular}{|c|c|c|c|}
\hline
Approach & Time Complexity & Space Complexity & FFT Type \\\hline
Original & $O(W_2H_2W_1H_1N^2)$ & $O(W_1H_1N^2)$ & N/A \\\hline
This Work & $O(W_2H_2W_1H_1N\log N)$ & $O(W_1H_1RNS)$ & 1-D\\\hline
\cite{mathieu2013fast} & $O(N^2W_0H_0\log W_0H_0)$ & $O(W_1H_1N^2)$ & 2-D \\\hline
\end{tabular}
\end{table}

\section{Experiments} \label{sec:exp}

\textbf{Dataset, Baseline \& Experiment Environment.} We evaluate our circulant structure-imposing approaches on two typical image classification datasets:  
CIFAR-10 \cite{krizhevsky2009learning} and ImageNet ILSVRC-2012 \cite{deng2009imagenet}. For each dataset, we take classical network models (ResNet \cite{he2016deep} for CIFAR-10 and AlexNet \cite{krizhevsky2012imagenet} for ImageNet) as the baseline models. The compressed circulant CNN models are generated by replacing convolutional layers of the baseline models with circulant convolutional layers. All models in this paper are trained using NVIDIA GeForce GTX 1080 GPUs and Intel(R) Xeon(R) CPU E5-2620 v4 @ 2.10GHz. 

\textbf{Selection of Training Strategy.} 
As presented in Section imposing CircConv,  
the circulant CNN model can be trained either from scratch or re-trained from a pre-trained non-circulant model. In our experiments we evaluate these two different training strategies on different datasets. Experimental results show that with the same compression configuration setting, the compressed circulant CNN models generated by these two training strategies have very similar test accuracies. Therefore in this paper we only report the results using training-from-scratch strategy.

\subsection{ResNet on CIFAR-10} \label{subsec:resnet}
In this experiment, ResNet-32 is selected as the baseline model due to its high accuracy and easiness of training. The training data is augmented by following the method in  \cite{simonyan2014very}: First pad each side of the image with four pixels and then apply $32\times32$ sized random  crops with horizontal flipping. The compressed ResNet-32 models are trained using stochastic gradient descent (SGD) optimizer with learning rate 0.1, momentum 0.9, batch size 64 and weight decay 0.0001.

\textbf{Model setting.} ResNet-32 consists of 15 \textit{convolutional blocks}, where each convolutional block contains two or three convolutional layers. Considering the number of possible compression configurations on different convolutional layers is very large, we choose to make the layers in the same block have the same compression ratio. In other words, a \textit{block-wise} compression strategy is adopted. Notice that because the first few convolutional layers of ResNet are very sensitive for compression \cite{he2016deep}, in this experiment we do not impose circulant structure to the convolutional layers in the 1st and 2nd blocks of ResNet-32. Besides, the 6th and 11th blocks are not compressed due to their  small weight tensor.

Table \ref{tbl:config_resnet} shows the detailed compression configurations for different convolutional blocks of ResNet-32. Here we explore 7 different compression configurations and then obtain 7 compressed models. For each compressed model, the compression ratios for its component convolutional blocks are listed in the row direction. Here each number $i$ in a specific compression configuration scheme indicates the compression ratio as $i$ for the convolutional layers in the corresponding convolutional block. When the block is associated with 1, that means the corresponding convolutional block is not compressed. Notice that due to the sensitivity of front blocks, for all the 7 models in Table \ref{tbl:config_resnet} the compression ratios of the front blocks are typically less than those of the later blocks.

\begin{table}[!tb]
\centering
\caption{
Compression Configurations. For the convolutional block with compression ratio $i$, all the convolutional layers in that block has the same compression ratio $i$.}
\label{tbl:config_resnet}
	
\begin{tabularx}{.7\columnwidth}{c| YYYYYYY}
\hline
Partitioned  & \multicolumn{7}{c}{Model ID} \\\cline{2-8}
Block ID & 1 & 2 & 3 & 4 & 5 & 6 & 7   \\ \hline
 1  & 1 & 1  & 1 & 1 & 1 & 1 & 1 \\\hline
 2  & 1 & 1  & 1 & 1 & 1 & 1 & 1 \\\hline
 3  & 1 & 1  & 2 & 2 & 2 & 4 & 4  \\\hline
 4  & 1 & 1  & 2 & 2 & 2 & 4 & 4  \\\hline
 5  & 1 & 1  & 2 & 2 & 2 & 4 & 4  \\\hline
 6  & 1 & 1  & 1 & 1 & 1 & 1 & 1 \\\hline
 7  & 1 & 1  & 2 & 2 & 4 & 4 & 8  \\\hline
 8  & 1 & 1  & 2 & 2 & 4 & 4 & 8  \\\hline
 9  & 1 & 1  & 2 & 2 & 4 & 4 & 8  \\\hline
 10 & 1 & 1  & 2 & 2 & 4 & 4 & 8  \\\hline
 11 & 1 & 1  & 1 & 1 & 1 & 1 & 1  \\\hline
 12 & 1 & 2  & 2 & 4 & 4 & 4 & 16 \\\hline
 13 & 1 & 2  & 2 & 4 & 4 & 4 & 16 \\\hline
 14 & 2 & 2  & 2 & 4 & 4 & 4 & 16 \\\hline
 15 & 2 & 2  & 2 & 4 & 4 & 4 & 16 \\ \hline
 \multicolumn{8}{c}{baseline \cite{he2016deep}:ResNet-32 without partitioning}\\\hline
\end{tabularx}

\end{table}

\begin{figure}[!ht]
\centering
\includegraphics[width = .7\columnwidth]{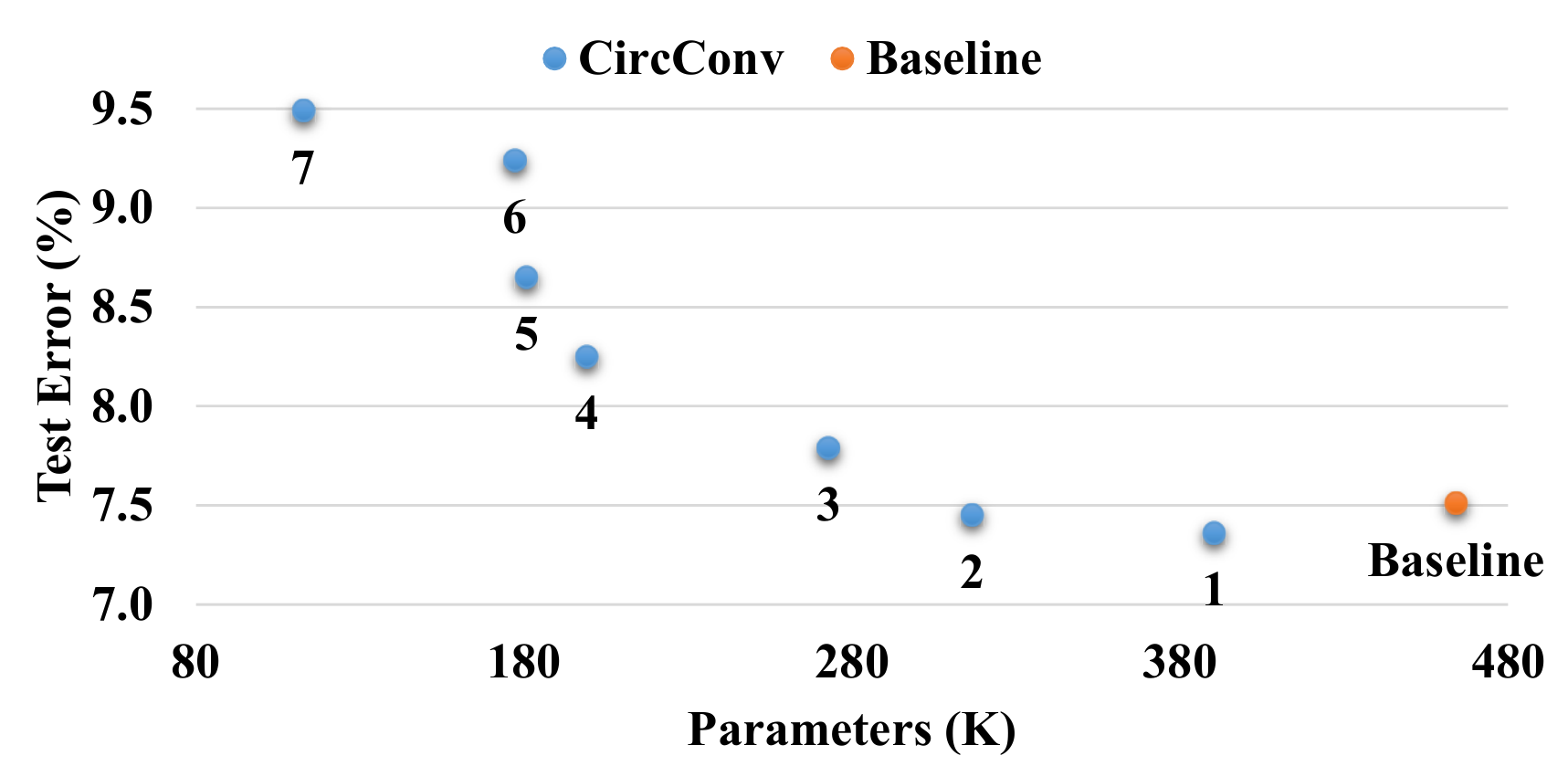}
\caption{ResNet-32 Test Error and Model Size. Use of circulant convolutional layer can bring half of parameters reduction with negligible test error increase. 
}
\label{fig:res-err}
\end{figure}

\textbf{Trade-off between accuracy and model size.}
Figure \ref{fig:res-err} shows the test error for 7 compressed models. It can be seen that model 1 even achieves slightly better performance with a smaller model size than the baseline. Moreover, model 2, 3 and 4 achieve around 50\% reduction in model size with negligible accuracy drop. 
With more aggressive compression configurations are selected  (such as model 5, 6 and 7), more reduction in model size can be further achieved with slight increase of test error. 

\begin{figure}[!ht]
\centering
\includegraphics[width = .7\columnwidth]{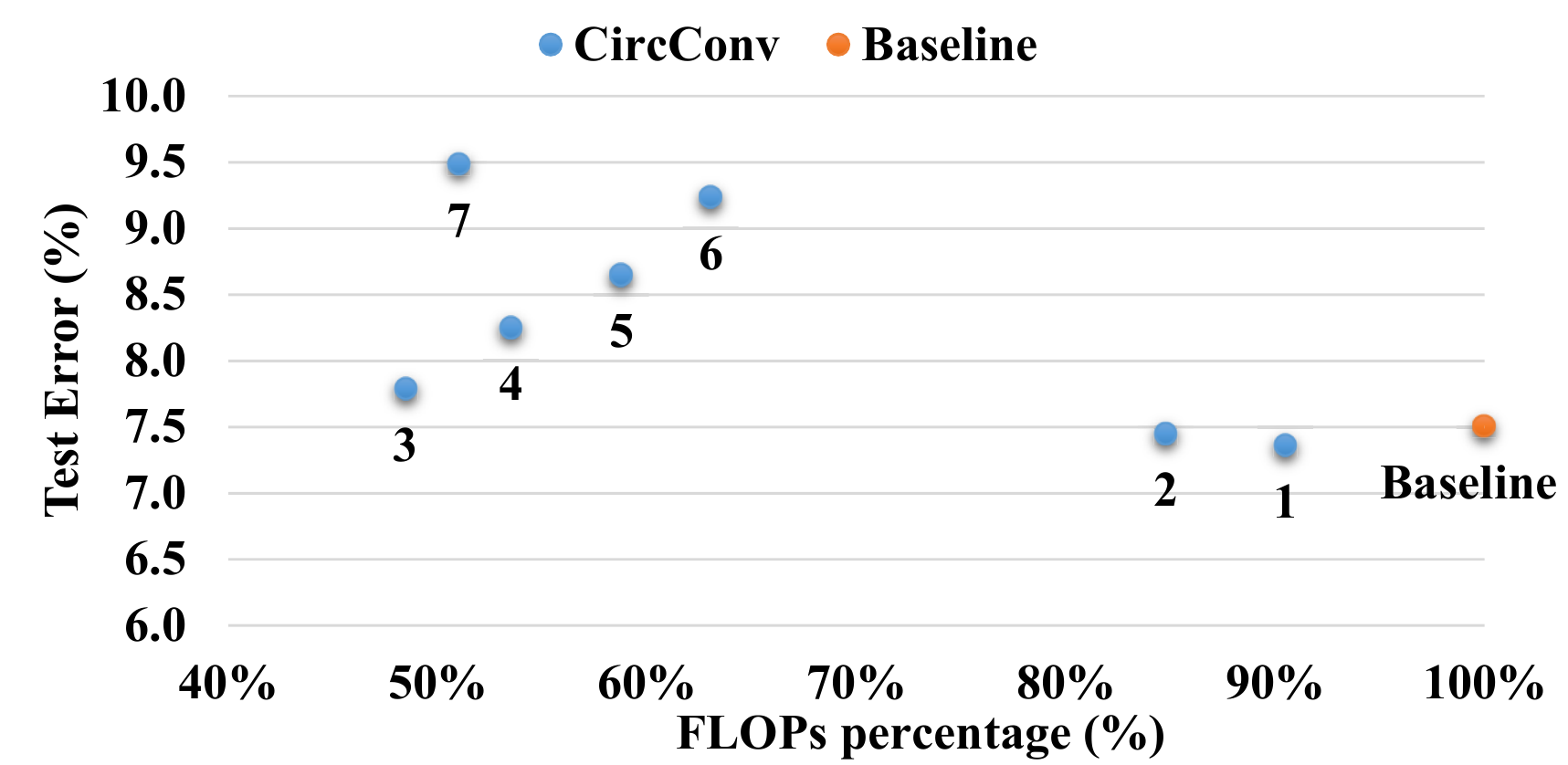}
\caption{ResNet-32 Test Error and Model Size. Use of circulant convolutional layer can bring half of FLOPs reduction with negligible test error increase.
}
\label{fig:res-flops}
\end{figure}

\textbf{Trade-off between accuracy and FLOPs.}
Our experiment also shows that the use of circulant convolutional layer helps reduce computational cost significantly. As shown in Figure \ref{fig:res-flops}, compressed model 1 and 2 achieve fewer FLOPs than baseline with the same or even less test error. For model 3, it can achieve 50\% reduction in FLOPs with negligible test error increase. An interesting discovery is that though model 4, 5 and 6 have more aggressive compression configurations than model 3, their corresponding reduction in FLOPs are less than what model 3 achieves. This is because the convolutional layers in the model 3 are mainly compressed with the factor of 2, which corresponds to 2-point FFT computation that only needs real number operations. 

\subsection{Wide ResNet on CIFAR-10}\label{subsec:wrn}

We also conduct the experiment on CIFAR-10 dataset using Wide ResNet \cite{zagoruyko2016wide}, which has better performance than conventional ResNet in term of test accuracy. In this experiment, the compressed Wide ResNet models are trained using SGD with learning rate 0.01, momentum 0.9, batch size 64 and weight decay 0.0005.

\textbf{Model settings.} To construct baseline Wide ResNet models, we take the same basic convolutional block structure in  \cite{zagoruyko2016wide} and set different numbers of convolutional blocks and widening parameters for different models. To achieve better performance, we add two more blocks to the convolutional blocks that are wider than $16\times k$, where $k$ is the inherent widening parameter of each block. Different from the experiment in 
ResNet experiment section, 
this experiment on Wide ResNet adopts very aggressive compression strategy: For one convolutional layer, if the numbers of input channels ($C_0$) and output channels ($C_2$) are the same, then the compression ratio for that layer is $i=C_0=C_2$; otherwise the convolutional layer is not compressed. We apply this compression strategy to five different Wide ResNet baselines and obtain five compressed Wide ResNet models. For each compressed model, it is labeled with two numbers ("$d$-$k$"), where $d$ and $k$ denote the number of convolutional layers ("depth") and widening parameter ("width"), respectively. These compressed models are compared with their corresponding baseline models as well as ResNet-110, which achieves the best performance on CIFAR-10 in \cite{he2016deep}. 

\begin{figure}{r}
    \begin{center}
    \includegraphics[width = .7\columnwidth]{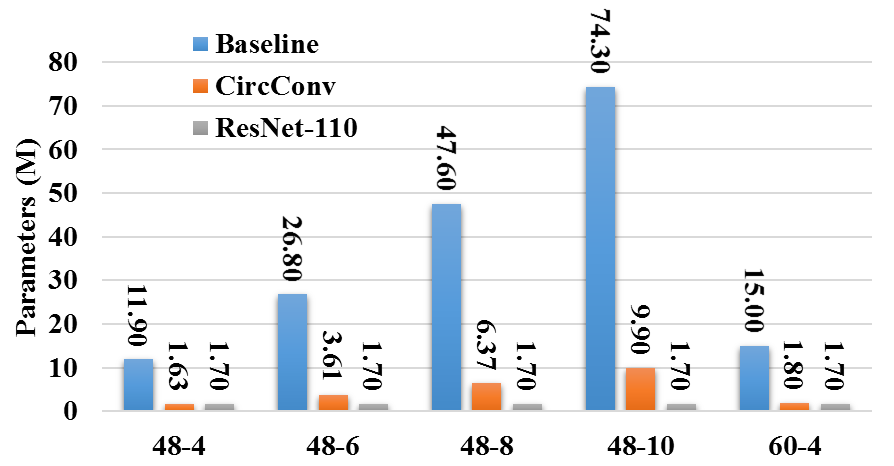}
    \end{center}
    \caption{Wide ResNet Model Size Reduction. Compared with baseline models, compressed models achieve similar model size as ResNet-110.  Compressed model named like "48-4" has 48 convolutional layers and widening parameter as 4.
    }
    \label{fig:wrn-param}
\end{figure}

\textbf{Model size reduction.} Figure \ref{fig:wrn-param} shows the number of parameters of Wide ResNet baselines and the corresponding compressed models after imposing circulant structure. It can be seen that the compressed models greatly reduce the model size. In particular, model "60-4" can achieve 8.35 times reduction in the  model size. Also, it can be seen that the numbers of parameters of Wide ResNet models are similar to the size of ResNet-110 after applying circulant convolutional layer. For instance, Model "48-4" has around 1.6M parameters which is less than 1.7M for ResNet-110.

\begin{figure}[h] 
\centering
\includegraphics[width = .7\columnwidth]{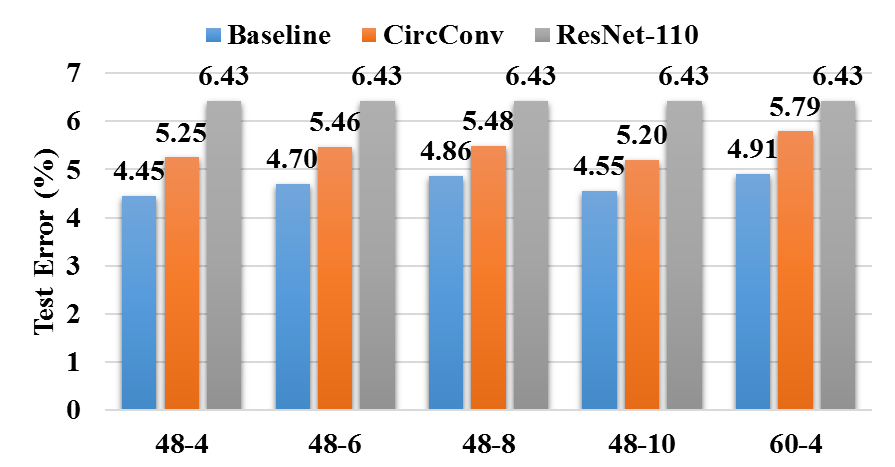}
\caption{Wide ResNet Test Error. Baseline models are different original Wide ResNets and they are compared with the corresponding compressed models and ResNet-110.
}
\label{fig:wrn-err}
\end{figure}

\textbf{Test error analysis.} Figure \ref{fig:wrn-err} shows test errors of baseline Wide ResNet models and the corresponding compressed models using circulant convolutional layer. It can be seen that all compressed models have slightly test error increase less than 1\%. In addition, compared with the state-of-the-art ResNet-110, all of the compressed models have around 1\% test error decrease.

\begin{figure}[h]
\centering
\includegraphics[width = .7\columnwidth]{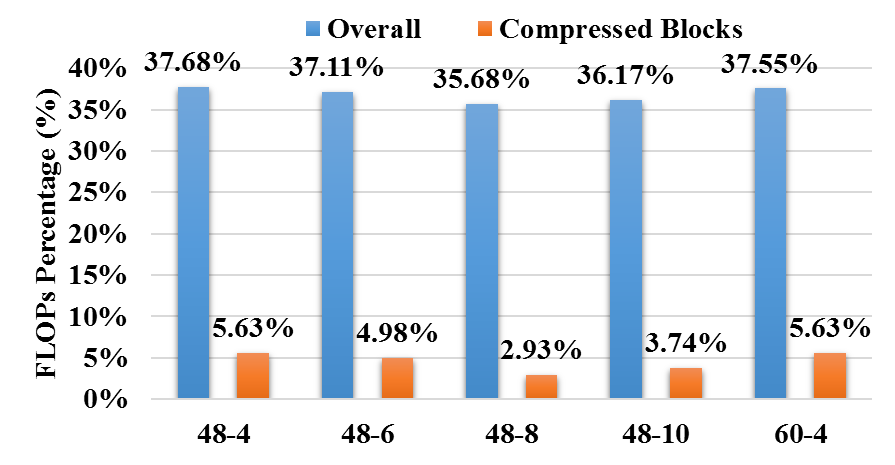}
\caption{Wide ResNet FLOPs. The overall FLOPs measure the FLOPs percentage of compressed models over corresponding baselines. We also list FLOPs percentage of compressed convolutional blocks over original blocks.}
\label{fig:wrn-flops}
\end{figure}

\textbf{Comparison with ResNet-110.} From Figure \ref{fig:wrn-err} we can see that all compressed Wide ResNet models have less test error than ResNet-110. Meanwhile, Figure \ref{fig:wrn-param} shows these compressed models have similar numbers of parameters as compared with ResNet-110, and model "48-4" has even fewer parameters. These results demonstrate that circulant structure-imposing approach can be useful in reducing model redundancy and holding less test error while maintaining similar model sizes.

\textbf{FLOPs reduction.} As shown in Figure \ref{fig:wrn-flops}, we measure the overall FLOPs reduction of Wide ResNet. It is found that the compressed Wide ResNet models can achieve significant reduction in FLOPs: all of them only require around 36\% FLOPs as compared to the corresponding baseline models. In addition, the FLOPs reduction for the compressed blocks are very significant. From Figure \ref{fig:wrn-flops} it can be seen that the FLOPs in the compressed blocks of all compressed models are only less than 6\% of the corresponding uncompressed blocks in the original Wide ResNet baseline models.

\subsection{AlexNet on ImageNet}
To test the effectiveness of the proposed circulant-imposing approach on large-scale datasets, we evaluate the performance of circulant CNNs on ImageNet (ILSVRC2012). Here the baseline model is AlexNet \cite{krizhevsky2012imagenet}. All training images are randomly distorted as suggested in \cite{szegedy2016rethinking}.  
We train our AlexNet models using RMSprop \cite{tieleman2012lecture} with learning rate 0.01, momentum 0.9, batch size 32 and decay 0.9.

\textbf{Model settings.} We explore three different compression configurations for the five convolutional layers in AlexNet. Table \ref{tbl:alexnet} listed the detailed compression configuration schemes by using notation "$a$-$b$-$c$-$d$-$e$". For instance, "1-2-2-2-2" means the first convolutional layer is not compressed, and the rest four layers are compressed with the factor of 2. By using these configurations, three compressed AlexNet models are generated and compared with original AlexNet baseline model. Also, since SSL in \cite{wen2016learning} is the state-of-the-art work that explores the relationship between accuracy and compressed model size for AlexNet, we also compare our circulant convolutional layer-based compressed AlexNet models with three SSL regularization-based compressed AlexNet models in \cite{wen2016learning}.

\begin{table}[t]
\centering
\caption{Comparison among  AlexNet models.}
\label{tbl:alexnet}

\begin{tabular}{c|c|c|c|c}
\hline
\makecell{AlexNet\\ Model} & \makecell{Compression\\ Configuration} & Test Error (\%) & Parameters (\%) & FLOPs(\%) \\\hline
\makecell{Original\\ (Baseline)} & N/A & 42.9 & 100 & 100 \\\hline
CircConv & 1-2-2-2-2 & \textbf{42.75} & 50.36  & \textbf{31.3} \\\hline
CircConv & 1-2-2-4-2 & 42.99 & 40.01 & \textbf{31.3} \\\hline
CircConv & 1-2-4-2-2 & 43.13 & 45.19 & \textbf{31.3} \\\hline
\cite{wen2016learning} & N/A & 42.75 & 51.20 & 39.0 \\\hline
\cite{wen2016learning} & N/A & 43.00 & 44.40 & 43.0 \\\hline
\cite{wen2016learning} & N/A & 43.25 & 42.30 & 45.0 \\\hline
\end{tabular}
\end{table}

\textbf{Test error analysis.} 
Table \ref{tbl:alexnet} shows the test errors of compressed AlexNet models by using circulant structure-imposing and SSL approaches. It can be seen that both these two approaches can render the compressed models with the similar test errors to the original AlexNet model. Among them, the circulant model with "1-2-2-2-2" compression configuration achieves the least test error.

\textbf{Model size reduction.} 
Table \ref{tbl:alexnet} 
shows the percentage of number of parameters of each model over original AlexNet model. It can be seen that circulant convolution-based models have similar numbers of parameters to SSL-based models. Among them the circulant convolution-based model with "1-2-2-4-2" compression configuration has the least number of parameters.

\textbf{FLOPs reduction.} 
Table \ref{tbl:alexnet}
shows the percentage of FLOPs of each model over the original AlexNet model. It can be found that all circulant convolution-based models require fewer FLOPs than the SSL regulation-based models. All the circulant convolution-based models have around 31\% FLOPs of original uncompressed AlexNet baseline.

\textbf{Overall comparison.} 
As shown in 
Table \ref{tbl:alexnet}
, circulant convolution-based models have similar accuracy to the state-of-the-art SSL models while maintaining similar number of parameters. Meanwhile, 
Table \ref{tbl:alexnet}
shows that circulant convolution-based models requires less FLOPs than the SSL models when targeting to the similar accuracy. Therefore, imposing circulant structure to convolutional layer is a very promising accuracy-retained approach to reduce both the space and computational costs.


\section{Conclusion} \label{sec:conclusion}
In this paper, we propose to impose the circulant structure to convolutional neural network. This structure-imposing approach leads to significant reduction in model size, FLOPs with negligible accuracy drop. Complexity analysis and experiments on different datasets and different network models demonstrate the effectiveness of the proposed approach.

\bibliographystyle{acm}

\bibliography{ref}

\end{document}